\documentclass{article}

\PassOptionsToPackage{numbers, compress}{natbib}

\usepackage[preprint,arxiv]{arxiv}

\usepackage{amsmath,amsfonts,bm}

\def\figref#1{figure~\ref{#1}}

\def\secref#1{section~\ref{#1}}

\def\eqref#1{equation~\ref{#1}}

\def\1{\bm{1}}

\DeclareMathAlphabet{\mathsfit}{\encodingdefault}{\sfdefault}{m}{sl}
\SetMathAlphabet{\mathsfit}{bold}{\encodingdefault}{\sfdefault}{bx}{n}

\newcommand{\KL}{D_{\mathrm{KL}}}

\usepackage{url}
\usepackage{marvosym}
\usepackage{xspace}
\usepackage{dsfont}
\usepackage{afterpage}

\usepackage{enumitem}
\usepackage{booktabs}
\usepackage{arydshln}
\usepackage{caption} 
\usepackage{multirow} 
\usepackage{enumitem}
\usepackage{colortbl} 
\usepackage{tabularx}
\usepackage{bbm}
\usepackage{algorithm}
\usepackage{algpseudocode}                  
\usepackage{setspace}                       
\usepackage{lipsum} 
\usepackage{subcaption} 
\usepackage{xcolor}
\usepackage{graphicx}
\usepackage[percent]{overpic}

\usepackage{wrapfig}
\usepackage{pifont}

\usepackage{mathtools} 
\usepackage{bm} 
\usepackage{soul} 
\usepackage{nicefrac}
\usepackage{csquotes}

\usepackage{duckuments}
\usepackage{xcolor}

\makeatletter
\newcommand*{\myfnsymbol}[1]{%
  \ensuremath{%
    \ifcase#1\or
      \text{\ding{41}}\or
      1\or
      2\or
      3\or
      4\or
      5\or
      6\or
      7\else
      8\fi
  }%
}
\def\@fnsymbol#1{\myfnsymbol{#1}}
\def\thempfootnote{\myfnsymbol{\c@mpfootnote}}
\makeatother

\definecolor{cornellred}{rgb}{0.7, 0.11, 0.11}
\definecolor{cadmiumgreen}{rgb}{0.0, 0.42, 0.24}
\definecolor{aliceblue}{rgb}{0.91, 0.94, 0.97}
\definecolor{darkblue}{rgb}{0.83, 0.89, 0.97}
\definecolor{Red7}{rgb}{0.941, 0.243, 0.243}
\definecolor{Green7}{RGB}{55, 178, 77}
\definecolor{Blue9}{rgb}{0.098,0.3,0.9}

\usepackage{pifont}

\sethlcolor{aliceblue}

\newcommand{\tabyes}{\ding{51}}
\newcommand{\tabno}{\ensuremath{\times}}
\definecolor{gaincolor}{RGB}{54,160,96}
\newcommand{\tabgain}[1]{\textcolor{gaincolor}{(+#1)}}
\newcommand{\tabbasegain}[1]{\textcolor{gray}{(+#1)}}
\newcommand{\tabhead}[1]{\begin{tabular}[c]{@{}>{\bfseries}c@{}}#1\end{tabular}}

\definecolor{citecolor}{HTML}{2980b9}
\definecolor{linkcolor}{HTML}{c0392b}
\definecolor{urlcolor}{RGB}{157,49, 251}

\usepackage{fontawesome5}
\newcommand{\mname}{Z-Reward\xspace}
\newcommand{\teaname}{GDSO\xspace}
\newcommand{\stuname}{RISD\xspace}

\usepackage{marvosym} 
\usepackage{amsmath} 
\usepackage[hidelinks,breaklinks=true,colorlinks,citecolor=citecolor,linkcolor=linkcolor,urlcolor=urlcolor]{hyperref}

\title{
Z-Reward: Beyond Scalar Rewards by\\Internalizing Reasoning into Score Distributions
}

\author{
\vspace{-39pt}\\
\makebox[\linewidth][c]{\textbf{Xin Jin}$^{1,2,*}$\hspace{2mm} \textbf{Huanqia Cai}$^{1,*,\dagger}$\hspace{2mm} \textbf{Zhen Li}$^{1}$\hspace{2mm} \textbf{Zechao Zhan}$^{1}$\hspace{2mm} \textbf{Dengyang Jiang}$^{1}$\hspace{2mm} \textbf{Aiming Hao}$^{1}$}\\[0.5mm]
\makebox[\linewidth][c]{\textbf{Yuming Jiang}$^{1}$\hspace{2mm} \textbf{Xiangpeng Yang}$^{1}$\hspace{2mm} \textbf{Chunle Guo}$^{2, }$\textsuperscript{\faEnvelope[regular]}\hspace{2mm} \textbf{Peng Gao}$^{1, }$\textsuperscript{\faEnvelope[regular]}\hspace{2mm} \textbf{Ming-Ming Cheng}$^{2}$\hspace{2mm} \textbf{Steven C.H. Hoi}$^{1}$}\\[1.5mm]
\makebox[\linewidth][c]{$^{1}$Z-Image Team, Alibaba Group \hspace{10mm} $^{2}$VCIP, CS, Nankai University}\\[2.4mm]
\makebox[\textwidth][c]{$^{*}$Equal contribution \hspace{10mm} $^{\dagger}$Project lead \hspace{10mm} \textsuperscript{\faEnvelope[regular]}Corresponding authors} \\[2mm]
\makebox[\linewidth][c]{\small{ \url{https://srameo.github.io/projects/z-reward/}}}
}

\begin{document}

\maketitle

\begin{abstract}
Reward models are central to text-to-image post-training, but visual preference is subjective and better represented as a distribution over rubric scores than as a deterministic scalar. Existing scalar, score-token, and pairwise reward models over-compress uncertainty and fine-grained score differences, while reasoning-based generative rewards provide stronger judgments but are costly to deploy and difficult to use as direct optimization signals. 
We propose \textbf{\mname}, a teacher-student reward modeling framework that decouples reasoning-heavy judgment from efficient reward deployment. The teacher is a large VLM that uses reasoning to infer rubric-aligned score distributions, and is trained with Group-wise Direct Score Optimization (\teaname), which combines policy-gradient rewards from distribution expectations with direct pointwise and pairwise supervision on score distributions and score gaps. The student is trained with Reasoning-Internalized Score Distillation (\stuname), which transfers the teacher's reasoning-conditioned score distribution into a compact VLM without requiring explicit reasoning chains at inference time. On our internally annotated evaluation set, the 27B \teaname teacher reaches 89.6\% human preference accuracy, outperforming SFT, RewardDance, and GRPO, while the 9B \stuname student reaches 88.6\%, outperforming the OPD baseline and closely matching the larger teacher. We further show that \mname can serve as a differentiable reward signal for text-to-image optimization, yielding a 41.3\% net human-preference improvement over SFT baseline. 
\end{abstract}

\section{Introduction}

\begin{figure}[H]
  \centering
  \begin{minipage}[t]{0.495\textwidth}
    \vspace{0pt}
    \includegraphics[width=\linewidth]{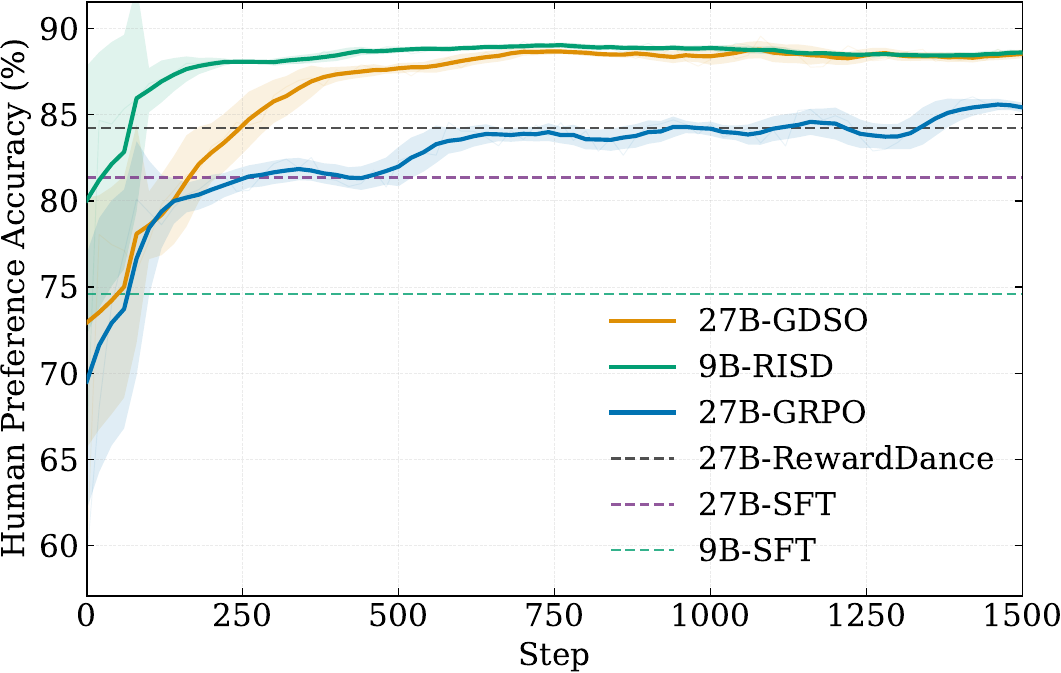}
  \end{minipage}
  \begin{minipage}[t]{0.48\textwidth}
    \vspace{0pt}
    \includegraphics[width=\linewidth]{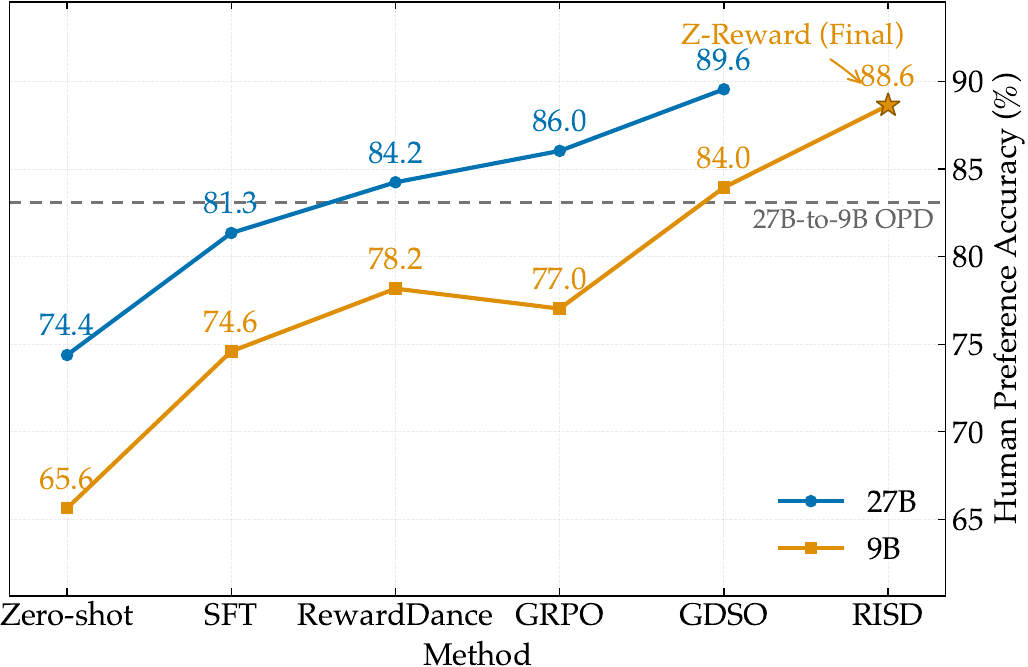}
  \end{minipage}
  \caption{Human preference accuracy for teacher optimization and student distillation. Left: accuracy curves over training steps show how reward-model performance evolves against SFT and RewardDance~\cite{wu2025rewarddancerewardscalingvisual} baselines. Right: final accuracy comparison shows that the 27B \teaname teacher outperforms SFT, RewardDance, and GRPO~\cite{deepseek-math}, while the 9B \stuname student reaches comparable performance to the larger teacher.}
  \label{fig:teaser-line-chart}
\end{figure}

\begin{table}[H]
  \centering
  \caption{Comparison of reward modeling paradigms for visual generation. Scalar or pairwise reward models are efficient but compress preference uncertainty, while reasoning-based generative reward models improve judgment quality at the cost of inference efficiency and direct differentiability. \mname separates these roles: the teacher uses reasoning to infer score distributions, and the student internalizes this ability for efficient direct scoring and gradient backpropagation.}
  \label{tab:reward-model-comparison}
  \resizebox{\textwidth}{!}{%
  \begin{tabular}{@{}c|c|c|c|c|c|c|c@{}}
  \toprule
  \textbf{Methods} &
  \textbf{\shortstack{Base\\Model}} &
  \textbf{\shortstack{Modeling\\Paradigm}} &
  \textbf{\shortstack{Training\\Strategy}} &
  \textbf{\shortstack{Scoring Based\\on Reasoning}} &
  \textbf{\shortstack{Score\\Distribution}} &
  \textbf{\shortstack{Support Gradient\\Backpropagation}} &
  \textbf{\shortstack{Inference\\Efficiency}} \\
  \midrule
  ImageReward~\cite{xu2023imagereward} & CLIP & Regressive & SFT & \tabno & \tabno & \tabyes & High \\
  PickScore~\cite{kirstain2023pickapicopendatasetuser} & CLIP & Regressive & SFT & \tabno & \tabno & \tabyes & High \\
  HPSv2~\cite{wu2023humanpreferencescorev2} & CLIP & Regressive & SFT & \tabno & \tabno & \tabyes & High \\
  VisionReward~\cite{xu2026visionrewardfinegrainedmultidimensionalhuman} & VLM & Regressive & SFT & \tabno & \tabno & \tabyes & High \\
  VideoAlign~\cite{liu2025improvingvideogenerationhuman} & VLM & Regressive & SFT & \tabno & \tabno & \tabyes & High \\
  HPSv3~\cite{ma2025hpsv3widespectrumhumanpreference} & VLM & Regressive & SFT & \tabno & \tabyes & \tabyes & High \\
  WorldPM~\cite{wang2025worldpmscalinghumanpreference} & VLM & Regressive & RL & \tabno & \tabno & \tabyes & High \\
  DeepSeek-GRM~\cite{liu2025inferencetimescalinggeneralistreward} & VLM & Generative & SFT & \tabyes & \tabyes & \tabno & Low \\
  Pairwise RM~\cite{liu2025pairjudgermperformbestofn} & VLM & Generative & RL & \tabno & \tabyes & \tabyes & High \\
  GenRM-CoT~\cite{zhang2025generativeverifiersrewardmodeling} & VLM & Generative & SFT & \tabyes & \tabyes & \tabyes & Low \\
  Edit-R1~\cite{guo2026leveraging} & VLM & Generative & RL & \tabyes & \tabno & \tabno & Low \\
  UnifiedReward~\cite{wang2026unifiedrewardmodelmultimodal} & VLM & Generative & SFT & \tabyes & \tabno & \tabno & Low \\
  RewardDance~\cite{wu2025rewarddancerewardscalingvisual} & VLM & Generative & SFT & \tabno & \tabyes & \tabyes & High \\\midrule
  \mname-Teacher & VLM & Generative & RL \& SFT & \tabyes & \tabyes & \tabyes & Low \\
  \mname-Student & VLM & Generative & Distillation & Internalized & \tabyes & \tabyes & High \\
  \bottomrule
  \end{tabular}%
  }
  \end{table}

Reward models are a key component of post-training, where they provide the preference signals used for model selection, data curation, and reward-guided optimization~\cite{Ouyang2022TrainingLM, Christiano2017DeepRL, xu2023imagereward, xu2026visionrewardfinegrainedmultidimensionalhuman, wu2025rewarddancerewardscalingvisual}. Unlike mathematics or coding rewards, however, visual preferences are inherently subjective: the same generated image can receive different judgments from different annotators, especially for aesthetics, realism, and fine-grained prompt alignment. Thus, human evaluation for visual generation is better viewed as a distribution of judgments rather than a deterministic scalar score~\cite{murray2012ava,talebi2018nima,wu2023qalign, you2025teachinglargelanguagemodels, ma2025hpsv3widespectrumhumanpreference,uma2021learning,davani2022dealing}.

As summarized in Table~\ref{tab:reward-model-comparison}, existing reward modeling paradigms each miss part of this requirement. Scalar, score-token, and pairwise reward models compress preference into a single value or comparison, which is efficient but discards annotator uncertainty and fine-grained differences among plausible scores~\cite{uma2021learning,davani2022dealing}. For example, two images may both collapse to score 4 under a discrete scoring scheme, even though one is slightly below the 4-point boundary and the other is slightly above it. Reasoning-based generative reward models can produce higher-quality judgments by leveraging world knowledge and explicit rationales~\cite{zheng2023judging,gu2024surveyllmasajudge,chen2024mllmjudge,chen2024mjbench}, but they are expensive at inference time and their textual reasoning or score outputs are less suitable for large-scale deployment and gradient-based optimization. Explicit distribution modeling~\cite{talebi2018nima,diaz2019soft,wen2023ordinal,you2025teachinglargelanguagemodels} can represent uncertainty more directly, but it typically relies on repeated annotations per sample, which is difficult to scale in production pipelines.

This creates a central tension for visual reward modeling: high-quality scoring requires reasoning and uncertainty awareness, while scalable post-training requires fast, direct, and differentiable reward signals. Building on knowledge distillation and sequence/rationale distillation~\cite{hinton2015distilling,kim2016sequence,hsieh2023distilling,gu2024minillm}, recent on-policy distillation (OPD) methods~\cite{lu2025onpolicy, zhao2026selfdistilledreasoneronpolicyselfdistillation, fu2026revisitingonpolicydistillationempirical,song2026surveyonpolicydistillationlarge, li2026rethinkingonpolicydistillationlarge, cui2026briefoverviewonpolicyselfdistillation} improve compact students by applying teacher-guided dense feedback to their own sampled reasoning trajectories, making on-policy reasoning distillation an increasingly important paradigm. For reward modeling, however, the deployment target is different: a reward model is expected to provide fast, stable, calibrated, and optimization-friendly scores, rather than expose long reasoning chains at inference time. Our key insight is that reward models do not need to reproduce how a teacher reasons; they need to reproduce how a reasoning teacher judges. Therefore, \mname resolves this tension by decoupling judgment quality from reward efficiency: instead of forcing the student to imitate the sequential process of reasoning, \mname allows the compact model to internalize the teacher's reasoning-conditioned judgment directly into score distributions.

\begin{figure}[t]
  \centering
  \begin{overpic}[width=\textwidth]{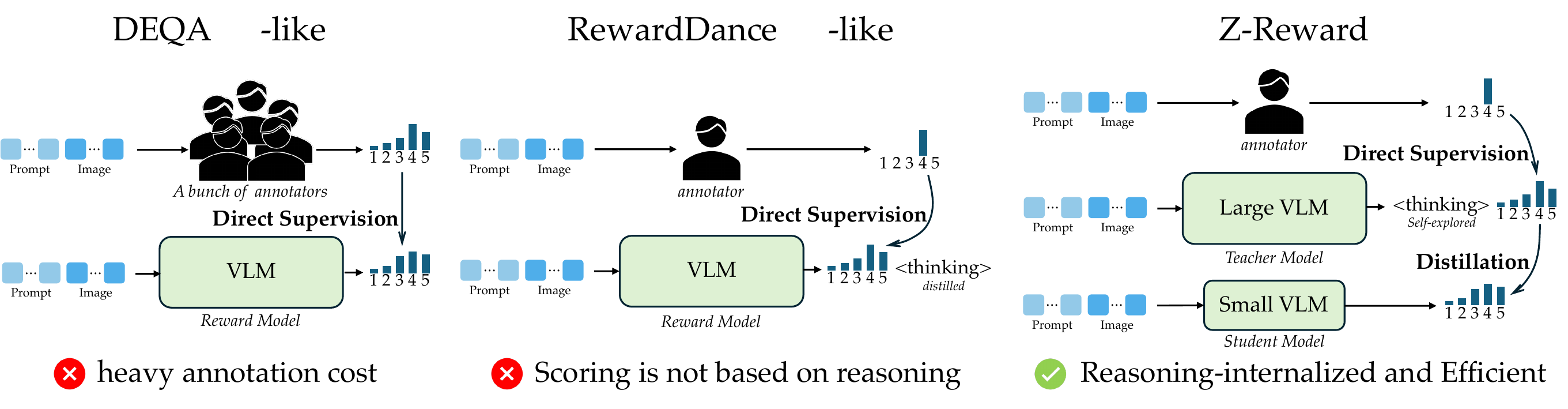}
      \put(13.8,23.55){\tiny\citep{you2025teachinglargelanguagemodels}}
      \put(50.0,23.55){\tiny\citep{wu2025rewarddancerewardscalingvisual}}
  \end{overpic}
  \caption{\textbf{Overview of \mname compared with existing distributional reward modeling paradigms.} Left: DEQA~\cite{you2025teachinglargelanguagemodels} rely on dense human score distributions for direct supervision, leading to heavy annotation cost.
Middle: RewardDance~\cite{wu2025rewarddancerewardscalingvisual} learn score distributions from direct supervision, but their scoring is not explicitly based on reasoning.
Right:  Our \mname first trains a reasoning-based large VLM teacher to infer calibrated score distributions, then distills this reasoning-enhanced distribution into a compact student that directly outputs scores without generating reasoning chains, enabling efficient deployment and gradient-based optimization.}
  \label{fig:teaser}
\end{figure}

We propose \mname, a teacher-student framework for reasoning-internalized score distributions, as illustrated in~\figref{fig:teaser}. The teacher is a large VLM that uses reasoning and world knowledge to infer a calibrated score distribution from scalable supervision. Here, reasoning is not used merely as an explanation artifact; it helps the teacher decompose visual evidence, apply rubric criteria, and allocate probability mass across neighboring score bins. The student is a compact reward model that internalizes this reasoning-enhanced distribution and directly predicts scores without generating reasoning chains at inference time, enabling efficient deployment and gradient backpropagation.

To train the teacher, we introduce \textbf{Group-wise Direct Score Optimization} (\teaname), which optimizes rewards computed from predicted score distributions and applies direct distribution-level supervision. Rather than requiring repeated human annotations to observe this distribution explicitly, we learn it as a latent, reasoning-conditioned distribution from scalable rubric-based supervision. To train the student, we further introduce \textbf{Reasoning-Internalized Score Distillation} (\stuname), which distills the teacher's  reasoning-conditioned score distribution into a small VLM without explicit reasoning tokens. Thus, the student does not imitate the teacher's reasoning text; it internalizes the distributional effect of that reasoning into direct scoring behavior. 

Empirically, this design leads to strong reward-modeling performance. As shown in Figure~\ref{fig:teaser-line-chart}, our 27B GDSO teacher reaches 89.6\% human preference accuracy, outperforming SFT, RewardDance-style supervision~\cite{wu2025rewarddancerewardscalingvisual}, and GRPO optimization~\cite{deepseek-math}.
More importantly, the 9B \stuname student reaches 88.6\%, outperforming the OPD student while closely matching the larger reasoning teacher, and remains efficient at inference time.
We further validate \mname as an optimizable reward signal by applying it to text-to-image post-training, where reward-guided optimization improves human preference over the SFT baseline.

Our contributions are summarized as follows:

\begin{itemize}
    \item We propose a reasoning-aware and uncertainty-aware teacher-student reward modeling framework that learns latent score distributions from scalable supervision.

    \item We introduce Group-wise Direct Score Optimization, which trains a reasoning-based VLM teacher by optimizing score distributions directly.

    \item We develop Reasoning-Internalized Score Distillation, allowing a compact student to internalize reasoning into efficient, direct, and differentiable scoring.

    \item Empirically, our 27B teacher substantially improves human preference accuracy over SFT, GRPO, and RewardDance, while the 9B student outperforms an OPD-based distillation baseline, nearly matches the teacher, and serves as an efficient reward signal for text-to-image optimization.

\end{itemize}

\section{Annotation and Datasets}
\label{sec:annotation_and_datasets}

\textbf{Annotation document.} We build the annotation document around four user- and production-critical dimensions: \textbf{Text--Image Alignment}, \textbf{Realism}, \textbf{Aesthetics}, and \textbf{Physical Plausibility}, following recent fine-grained and multi-dimensional human-feedback settings for text-to-image generation~\cite{liang2024rich,zhang2024learning_mps}. Each dimension is scored with a five-level rubric that specifies how different error patterns should affect the score, rather than relying only on abstract terms such as ``minor'' or ``major.'' Although the rubric is organized around five integer-level anchors, final annotations are recorded on a nine-level half-point scale, i.e., 
$\hat{s}\in\{1.0,1.5,\ldots,5.0\}$. 
This half-point annotation scheme allows annotators to capture fine-grained quality differences between samples that fall into the same coarse rubric bin. For example, a score of 4 corresponds to one or two subtle defects, while a score of 3 indicates more salient subject-level errors or clearly visible quality degradation. To make these criteria operational, each score bin in each dimension is paired with 15--20 annotated examples, allowing annotators to calibrate new samples through a nearest-neighbor-style comparison. The document is updated throughout annotation by adding newly discovered corner cases and replacing less representative examples with more typical ones.

\textbf{Data for annotation and evaluation.} Our annotation prompts come from three sources: 1) internal captions rewritten as generation prompts; 2) real-world prompts from users or community usage; and 3) concepts sampled from our topology, composed, and LLM-expanded into diverse prompts, covering compositional phenomena emphasized by T2I evaluation benchmarks~\cite{hu2023tifa,ghosh2023geneval,huang2023t2icompbench,saharia2022photorealistic}. For evaluation, we construct a held-out test set with multiple annotations per sample. To compute the ground-truth score distribution, we drop the highest and lowest scores before aggregation to reduce outliers and stabilize preference estimates~\cite{uma2021learning,davani2022dealing}.

\begin{figure}[H]
  \centering
  \begin{overpic}[width=\textwidth]{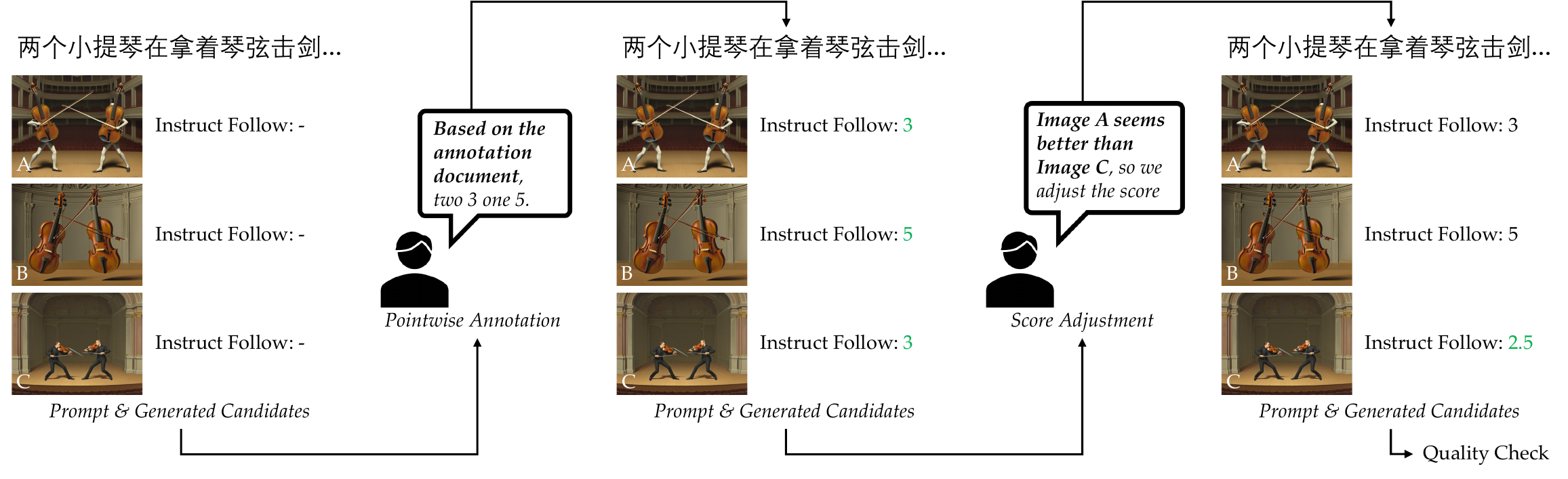}
  \end{overpic}
  \caption{\textbf{Annotation workflow.} For each prompt, annotators 1) assign pointwise scores to generated candidates according to the annotation document, 2) compare candidates under the same prompt to refine scores within the same coarse bin, and 3) send the resulting annotations to quality check before they are admitted into the training set.}
  \label{fig:annotation}
\end{figure}

\textbf{Annotation workflow.} As shown in~\figref{fig:annotation}, annotators 1) assign pointwise scores to generated candidates according to the rubric and example document, 2) compare candidates under the same prompt and within the same coarse score bin to shift distinguishable samples by $\pm 0.5$, and 3) submit the results to quality-control annotators for final review. Only data from annotators whose audited accuracy exceeds a preset threshold is admitted into the training set.

\textbf{The risk of context mismatch.} This process exposes two context mismatches between annotators and reward models: 1) the full annotation document is too long to place into a reward model's context, since four dimensions, five score bins, and 15 images per bin already require approximately $4 \times 5 \times 15 \times 1024 = 307{,}200$ image tokens before counting textual instructions; and 2) annotators can compare same-prompt candidates during score adjustment, while a deployable pointwise reward model only observes one text--image pair at a time. These mismatches motivate direct supervision on scores and score distributions, so the model can learn human-calibrated scoring behavior without relying on the full annotation context at inference time.

\section{Method}

The annotation process described above provides calibrated human scores, but the calibration context available to annotators cannot be directly supplied to a deployable reward model. Annotators can consult a long, evolving document and compare candidates under the same prompt, whereas a pointwise reward model must usually judge one text--image pair at a time. We therefore decouple reward-model training into a reasoning-intensive teacher stage and an efficient student stage. The teacher learns a reliable reasoning-augmented score distribution, while the student internalizes this distribution into a compact model for direct scoring and gradient backpropagation.

Given a prompt $p$, an image $I$, and a reward dimension $d \in \mathcal{D}$, the teacher first generates a reasoning trace $\rho$ and then predicts a distribution over rubric-aligned score bins $s \in \mathcal{S}$:
\begin{equation}
q_\theta(s \mid p,I,d,\rho), \quad s \in \mathcal{S}.
\label{eq:method-score-distribution}
\end{equation}

We treat $q_\theta$ as a predictive distribution over rubric-aligned score bins, rather than as a directly observed empirical annotator distribution for each training sample. Since scalable annotation usually provides one rubric-calibrated score per text--image pair, the distribution is learned implicitly from large-scale score supervision, same-prompt score-gap constraints, and the teacher's reasoning-conditioned score-token probabilities. This follows the ordinal and soft-label view that neighboring score bins should carry structured uncertainty rather than being treated as unrelated classes~\cite{diaz2019soft,wen2023ordinal}. Thus, $q_\theta$ captures the model's uncertainty over plausible neighboring bins while its expectation is calibrated to human rubric scores.

We decode this distribution from the teacher's score tokens using a Q-Align-style~\cite{wu2023qalign} score decoder. The expected scalar score is then obtained from the decoded distribution:
\begin{equation}
\mu_\theta(p,I,d,\rho) = \sum_{s \in \mathcal{S}} s \, q_\theta(s \mid p,I,d,\rho).
\label{eq:method-expected-score}
\end{equation}
These reasoning traces help decompose visual evidence, apply fine-grained rubric rules, and handle corner cases that would otherwise be compressed into one-hot labels. However, explicit reasoning is expensive and unsuitable for deployment. We therefore train a compact student to predict the teacher's reasoning-conditioned distribution directly as $q_\phi(s \mid p,I,d)$, without producing reasoning chains at inference time. The following subsections describe teacher optimization and student distillation separately.

\subsection{Training Teacher Model via Group-wise Direct Score Optimization}

\begin{algorithm}[t]
\caption{Iterative Group-wise Direct Score Optimization (\teaname)}
\label{alg:gdso}
\begin{algorithmic}[1]
  \Statex \textbf{Input} initial teacher policy $\pi_{\theta_{\mathrm{init}}}$; annotated preference data $\mathcal{D}_{\mathrm{ann}}$
  \Statex \hspace{\algorithmicindent} score bins $\mathcal{S}$; score range $[S_{\min},S_{\max}]$; group size $G$
  \Statex \hspace{\algorithmicindent} hyperparameters $\beta,\epsilon,\lambda_{\mathrm{pt}},\lambda_{\mathrm{pw}},\alpha_{\mathrm{pt}},\alpha_{\mathrm{pw}}$
  \State policy model $\pi_\theta \gets \pi_{\theta_{\mathrm{init}}}$
  \For{iteration $=1,\ldots,N$}
    \State reference model $\pi_{\mathrm{ref}} \gets \pi_\theta$
    \For{step $=1,\ldots,M$}
      \State Sample a batch $\mathcal{B}$ from $\mathcal{D}_{\mathrm{ann}}$
      \State Update the old policy model $\pi_{\theta_{\mathrm{old}}}\gets\pi_\theta$
      \ForAll{$(p,I_w,I_l,\hat{s}_w,\hat{s}_l,d)\in\mathcal{B}$}
        \State Set $x_w=(p,I_w,d)$ and $x_l=(p,I_l,d)$
        \For{$j\in\{w,l\}$}
          \State Sample $G$ outputs $\{o_{j,i}\}_{i=1}^{G}\sim\pi_{\theta_{\mathrm{old}}}(\cdot\mid x_j)$
          \State Decode $\rho_{j,i}$, $q_{j,i}(s)$, and $\mu_{j,i}$ from each $o_{j,i}$
        \EndFor
        \State Compute rewards $r^{\mathrm{pt}}_{j,i}$, $r^{\mathrm{pw}}_{j,i}$, and $r_{j,i}$ using Eqs.~(\ref{eq:pointwise-reward}), ~(\ref{eq:pairwise-reward}), and ~(\ref{eq:gdso-reward})
        \State Compute group-relative advantages $A_{j,i}$ using Eq.~(\ref{eq:grpo-advantage})
        \State Compute direct losses $\mathcal{L}^{\mathrm{pt}}_{\mathrm{CE}}$ and $\mathcal{L}^{\mathrm{pw}}$ using Eqs.~(\ref{eq:pointwise-ce}) and~(\ref{eq:pairwise-loss})
      \EndFor
      \For{GDSO update $=1,\ldots,K_{\mathrm{gdso}}$}
        \State Update $\pi_\theta$ by minimizing $\mathcal{L}_{\mathrm{GDSO}}$ in Eq.~(\ref{eq:gdso-objective})
      \EndFor
    \EndFor
  \EndFor
  \Statex \textbf{Output} optimized teacher policy $\pi_\theta$
\end{algorithmic}
\end{algorithm}

We begin from Group Relative Policy Optimization (GRPO)~\cite{deepseek-math}, which samples a group of responses for the same input and optimizes the policy using group-normalized advantages. Given an input $x$, sampled responses $\{o_i\}_{i=1}^{G}$, and rewards $\{r_i\}_{i=1}^{G}$, the advantage of each response is normalized within the group:
\begin{equation}
A_i = \frac{r_i - \operatorname{mean}(\{r_k\}_{k=1}^{G})}
{\operatorname{std}(\{r_k\}_{k=1}^{G}) + \epsilon}.
\label{eq:grpo-advantage}
\end{equation}
The GRPO objective combines a policy-gradient term with KL regularization to a reference policy:
\begin{equation}
\mathcal{L}_{\mathrm{GRPO}}
=
-\mathbb{E}_{o_i \sim \pi_\theta(\cdot \mid x)}
\left[
A_i \sum_t \log \pi_\theta(o_{i,t} \mid o_{i,<t},x)
- \beta \KL\!\left(\pi_\theta(\cdot \mid x)\,\|\,\pi_{\mathrm{ref}}(\cdot \mid x)\right)
\right].
\label{eq:grpo-objective}
\end{equation}

GRPO alone treats the parsed score as a scalar reward, which can be slow to calibrate under the context mismatch discussed above. We therefore introduce \textbf{Group-wise Direct Score Optimization} (\teaname), which augments policy-gradient optimization with direct supervised gradients on score distributions and score gaps. Each training instance contains a winning sample $x_w=(p,I_w,d)$ and a losing sample $x_l=(p,I_l,d)$ with ground-truth rubric scores $\hat{s}_w$ and $\hat{s}_l$. For each side $j \in \{w,l\}$, the teacher samples $G$ reasoning-and-score outputs $o_{j,i}=(\rho_{j,i},a_{j,i})$:
\begin{equation}
o_{j,i} \sim \pi_\theta(\cdot \mid x_j), \quad i=1,\ldots,G.
\label{eq:gdso-sampling}
\end{equation}
Following the Q-Align-style score decoder introduced above, each output is converted into a predicted score distribution $q_{j,i}(s)=q_\theta(s\mid x_j,\rho_{j,i})$ and an expected score:
\begin{equation}
\mu_{j,i} = \sum_{s \in \mathcal{S}} s \, q_{j,i}(s).
\label{eq:parsed-score}
\end{equation}

Unlike most generative reward methods that parse a single textual score from the model response and then treat the parsed value as the reward, \teaname treats the decoded score distribution as the optimization target. Rewards are computed from the expectation of $q_{j,i}$, while direct losses supervise the score-bin distribution and its induced score gaps.

Importantly, \teaname is group-wise not only in the GRPO-style advantage normalization, but also in its direct score supervision. For each candidate, pointwise supervision is applied to all $G$ sampled score distributions in its group. For each same-prompt candidate pair, pairwise supervision is applied across all $G \times G$ cross-side sampled output pairs. Thus, the sampled group provides multiple reasoning-conditioned distributional views for both policy optimization and direct score calibration.

\textbf{Pointwise score supervision.}
For each sampled output, the pointwise reward measures how close its decoded expected score is to the annotated score:
\begin{equation}
r^{\mathrm{pt}}_{j,i}
=
1 -
\frac{\left|\mu_{j,i}-\hat{s}_j\right|}{S_{\max}-S_{\min}}.
\label{eq:pointwise-reward}
\end{equation}
As a policy reward, this term favors outputs whose decoded scores stay close to the rubric-calibrated human score, encouraging the teacher to learn an absolute score scale rather than only a relative preference direction. In addition to using this value as a policy-gradient reward, we directly supervise the decoded score distribution with a cross-entropy loss:
\begin{equation}
\mathcal{L}^{\mathrm{pt}}_{\mathrm{CE}}
=
-\frac{1}{2G}
\sum_{j \in \{w,l\}}\sum_{i=1}^{G}
\log q_{j,i}(\hat{s}_j).
\label{eq:pointwise-ce}
\end{equation}
This supervised loss anchors the score-bin probability to the annotated bin, boosting score-scale convergence so the policy does not need to discover the scoring convention only through sampled rewards. The soft distribution around ordinal score bins also provides a more informative target than a one-hot nominal label~\cite{diaz2019soft,wen2023ordinal}.

\textbf{Pairwise score-gap supervision.}
Pointwise supervision calibrates absolute scores, while pairwise supervision preserves the relative score gap between samples under the same prompt. Let $\bar{w}=l$, $\bar{l}=w$, and
\begin{equation}
\Delta \hat{s}_{j,\bar{j}} = \hat{s}_j - \hat{s}_{\bar{j}}.
\label{eq:score-gap}
\end{equation}
For a sampled output $o_{j,i}$, the pairwise reward compares its score gap to every sampled output from the opposite side:
\begin{equation}
r^{\mathrm{pw}}_{j,i}
=
1 -
\frac{1}{G(S_{\max}-S_{\min})}
\sum_{k=1}^{G}
\left|
(\mu_{j,i}-\mu_{\bar{j},k})-\Delta \hat{s}_{j,\bar{j}}
\right|.
\label{eq:pairwise-reward}
\end{equation}
As a policy reward, this term favors outputs whose score differences match the annotated gap across same-prompt candidates, encouraging the teacher to learn both the preference direction and the magnitude of visual quality differences. The corresponding direct pairwise loss is
\begin{equation}
\mathcal{L}^{\mathrm{pw}}
=
\frac{1}{2G^2(S_{\max}-S_{\min})}
\sum_{j \in \{w,l\}}\sum_{i=1}^{G}\sum_{k=1}^{G}
\left|
(\mu_{j,i}-\mu_{\bar{j},k})-\Delta \hat{s}_{j,\bar{j}}
\right|.
\label{eq:pairwise-loss}
\end{equation}
This supervised gap loss boosts within-prompt discrimination while keeping score margins calibrated to the annotation scale, rather than allowing the policy to separate pairs with arbitrary large margins.

We use score-gap supervision instead of a Bradley--Terry objective~\cite{Bradley1952RANKAO} or a binary preference-optimization objective such as DPO~\cite{rafailov2023direct}. A Bradley--Terry model estimates the preference probability as
\begin{equation}
P(x_w \succ x_l)
=
\sigma(\mu_w-\mu_l),
\quad
\mathcal{L}_{\mathrm{BT}}
=
-\log \sigma(\mu_w-\mu_l).
\label{eq:bt-loss}
\end{equation}
This objective only requires the winner score to exceed the loser score and can keep enlarging the margin, even when both absolute scores should stay close to the annotation rubric. In contrast, our pairwise term matches the annotated score gap, making it consistent with pointwise calibration.

The final GDSO reward combines pointwise and pairwise rewards:
\begin{equation}
r_{j,i}
=
\lambda_{\mathrm{pt}} r^{\mathrm{pt}}_{j,i}
+
\lambda_{\mathrm{pw}} r^{\mathrm{pw}}_{j,i}.
\label{eq:gdso-reward}
\end{equation}
The overall teacher-training objective is
\begin{equation}
\mathcal{L}_{\mathrm{GDSO}}
=
\mathcal{L}_{\mathrm{GRPO}}(\{r_{j,i}\})
+
\alpha_{\mathrm{pt}}\mathcal{L}^{\mathrm{pt}}_{\mathrm{CE}}
+
\alpha_{\mathrm{pw}}\mathcal{L}^{\mathrm{pw}}.
\label{eq:gdso-objective}
\end{equation}
Thus, \teaname does not rely on policy-gradient reward alone: the score distribution and score gap both receive supervised gradients, which accelerates score-scale calibration and score-distribution convergence. Algorithm~\ref{alg:gdso} summarizes \teaname in a GRPO-style iterative procedure.

\subsection{Teaching Student Model via Reasoning-Internalized Score Distillation}

After teacher training, the large reasoning model generates a reasoning trace $\rho_T$ and produces a calibrated distribution $q_T(s \mid p,I,d,\rho_T)$ for each text--image pair and reward dimension. The student model $q_\phi(s \mid p,I,d)$ is trained to predict this distribution directly, without generating the teacher's reasoning chain. Unlike sequence-level or rationale distillation, which transfers generated trajectories or explanatory traces~\cite{kim2016sequence,hsieh2023distilling}, \stuname uses the teacher's reasoning-conditioned score distribution as a soft target in the spirit of knowledge distillation~\cite{hinton2015distilling}. We distill the teacher distribution with a KL loss:
\begin{equation}
\mathcal{L}_{\mathrm{RISD}}
=
\mathbb{E}_{(p,I,d)}
\left[
\KL\!\left(
q_T(s \mid p,I,d,\rho_T)\,\|\,q_\phi(s \mid p,I,d)
\right)
\right].
\label{eq:risd-loss}
\end{equation}
The student score used for deployment is the expectation of the distilled distribution:
\begin{equation}
\mu_\phi(p,I,d)
=
\sum_{s \in \mathcal{S}} s \, q_\phi(s \mid p,I,d).
\label{eq:student-expected-score}
\end{equation}
Algorithm~\ref{alg:risd} summarizes the \stuname distillation procedure. By internalizing the teacher's reasoning-conditioned distribution, the student preserves much of the teacher's reward-modeling ability while avoiding explicit reasoning at inference time. This yields a compact reward model that supports efficient pointwise scoring and differentiable reward-guided optimization.

\begin{algorithm}[t]
  \caption{Reasoning-Internalized Score Distillation (\stuname)}
  \label{alg:risd}
  \begin{algorithmic}[1]
    \Statex \textbf{Input} trained teacher policy $\pi_{\theta_T}$; distillation data $\mathcal{D}_{\mathrm{dist}}=\{(p,I,d)\}$; initial student $q_{\phi_{\mathrm{init}}}$
    \Statex \hspace{\algorithmicindent} score bins $\mathcal{S}$; hyperparameters for student optimization
    \State student reward model $q_\phi \gets q_{\phi_{\mathrm{init}}}$
    \State Freeze the teacher policy $\pi_{\theta_T}$
    \For{step $=1,\ldots,M_{\mathrm{dist}}$}
      \State Sample a batch $\mathcal{B}$ from $\mathcal{D}_{\mathrm{dist}}$
      \ForAll{$(p,I,d)\in\mathcal{B}$}
        \State Query $\pi_{\theta_T}$ to generate reasoning $\rho_T$ and decode $q_T(s\mid p,I,d,\rho_T)$
        \State Predict student distribution $q_\phi(s\mid p,I,d)$ without reasoning tokens
      \EndFor
      \State Update $q_\phi$ by minimizing $\mathcal{L}_{\mathrm{RISD}}$ in Eq.~(\ref{eq:risd-loss})
    \EndFor
    \Statex \textbf{Output} deployable student $q_\phi(s\mid p,I,d)$ and score $\mu_\phi(p,I,d)$ in Eq.~(\ref{eq:student-expected-score})
  \end{algorithmic}
\end{algorithm}

\section{Experiment}

\subsection{Experimental Setup}

\textbf{Model choices.} We instantiate the teacher with Qwen3.5-27B~\cite{qwen3.5} and the student with Qwen3.5-9B~\cite{qwen3.5}. This setting follows the design goal of \mname: the larger teacher provides stronger reasoning and score-distribution estimation, while the smaller student is used to test whether the reasoning-conditioned distribution can be internalized into an efficient reward model.

\textbf{Evaluation data and metrics.} We evaluate all reward models on the held-out test set from~\secref{sec:annotation_and_datasets}. We report PLCC and SRCC for score calibration, and human preference accuracy (HPA) and margin HPA for preference ranking. Margin HPA uses only pairs with a human score gap above 0.5.

For preference-ranking evaluation, we compute HPA over same-prompt candidate pairs with non-tied human scores. Given a pair $(I_a,I_b)$, HPA is defined as
\begin{equation}
\mathrm{HPA}
=
\frac{1}{|\mathcal{P}|}
\sum_{(a,b)\in\mathcal{P}}
\mathbbm{1}
\left[
(\mu_a-\mu_b)(\hat{s}_a-\hat{s}_b) > 0
\right],
\label{eq:hpa}
\end{equation}
where $\mathcal{P}$ denotes all evaluated pairs with $\hat{s}_a \neq \hat{s}_b$, $\hat{s}$ is the aggregated human score, and $\mu$ is the reward model's predicted expected score. Margin HPA is computed on the subset satisfying $|\hat{s}_a-\hat{s}_b|>0.5$.

\textbf{Compared methods.} We first include a zero-shot baseline, which evaluates the base model before SFT using only the scoring system prompt. We then compare against standard SFT,  which fine-tunes the same backbone on annotated score outputs without reasoning chains and serves as the no-reasoning baseline for teacher-side comparison; RewardDance~\cite{wu2025rewarddancerewardscalingvisual}, which uses post-hoc pseudo reasoning chains distilled from Qwen-3.6-Max; and GRPO, which computes rewards from the mean of the predicted score distribution as the final output score. We also evaluate our \teaname, which directly optimizes score distributions with pointwise and pairwise supervision. For a clean comparison of reinforcement learning strategies, all GRPO and \teaname runs start from the base model; their reasoning and scoring behaviors are driven only by the system prompt and learned through pure self-exploration. These comparisons are conducted for both 27B and 9B models. For the 9B setting, we additionally evaluate \stuname, which distills the 27B reasoning-based score distribution into the smaller student model.

\begin{table*}[t]
\centering
\caption{\textbf{Reward-model evaluation on the internally annotated test set.} We compare score calibration and preference-ranking quality using PLCC, SRCC, human preference accuracy, and margin human preference accuracy. Values in parentheses report absolute gains over the zero-shot baseline within the same model size. The \textbf{best} and \underline{second-best} results within each model size are highlighted. Margin human preference accuracy is computed only on pairs whose human score gap is larger than 0.5.}
\resizebox{\textwidth}{!}{%
\begin{tabular}{l c c c c c}
\toprule
\tabhead{Method} &
\tabhead{Scoring Based\\on Reasoning} &
\tabhead{PLCC} &
\tabhead{SRCC} &
\tabhead{Human Preference\\Accuracy} &
\tabhead{Margin Human\\Preference Accuracy} \\
\midrule
\multicolumn{6}{l}{\textcolor{gray}{\emph{Qwen3.5-27B}}} \\
Zero-shot & \tabno & 0.6301 \tabbasegain{.0000} & 0.5816 \tabbasegain{.0000} & 0.7438 \tabbasegain{.0000} & 0.9538 \tabbasegain{.0000} \\
SFT & \tabno & 0.6458 \tabgain{.0157} & 0.5914 \tabgain{.0098} & 0.8135 \tabgain{.0697} & 0.9644 \tabgain{.0106} \\
RewardDance & \tabno & 0.6667 \tabgain{.0366} & 0.6207 \tabgain{.0391} & 0.8425 \tabgain{.0987} & 0.9706 \tabgain{.0168} \\
GRPO & \tabyes & \underline{0.7200} \tabgain{.0899} & \underline{0.6832} \tabgain{.1016} & \underline{0.8604} \tabgain{.1166} & \underline{0.9827} \tabgain{.0289} \\
\rowcolor[RGB]{240,230,245}
\teaname & \tabyes & \textbf{0.7620} \tabgain{.1319} & \textbf{0.7132} \tabgain{.1316} & \textbf{0.8956} \tabgain{.1518} & \textbf{0.9885} \tabgain{.0347} \\
\arrayrulecolor{black!40}\midrule
\multicolumn{6}{l}{\textcolor{gray}{\emph{Qwen3.5-9B}}} \\
Zero-shot & \tabno & 0.3411 \tabbasegain{.0000} & 0.3167 \tabbasegain{.0000} & 0.6563 \tabbasegain{.0000} & 0.7501 \tabbasegain{.0000} \\
SFT & \tabno & 0.5296 \tabgain{.1885} & 0.4942 \tabgain{.1775} & 0.7459 \tabgain{.0896} & 0.8401 \tabgain{.0900} \\
RewardDance & \tabno & 0.5182 \tabgain{.1771} & 0.4338 \tabgain{.1171} & 0.7817 \tabgain{.1254} & 0.8972 \tabgain{.1471} \\
GRPO & \tabyes & 0.5340 \tabgain{.1929} & 0.5072 \tabgain{.1905} & 0.7703 \tabgain{.1140} & 0.9076 \tabgain{.1575} \\
\teaname & \tabyes & \underline{0.6341} \tabgain{.2930} & \underline{0.5665} \tabgain{.2498} & \underline{0.8395} \tabgain{.1832} & \underline{0.9599} \tabgain{.2098} \\
\rowcolor[RGB]{240,230,245}
\stuname & Internalized & \textbf{0.7391} \tabgain{.3980} & \textbf{0.6882} \tabgain{.3715} & \textbf{0.8864} \tabgain{.2301} & \textbf{0.9801} \tabgain{.2300} \\
\arrayrulecolor{black}\bottomrule
\end{tabular}%
}
\label{tab:reward_model_evaluation}
\end{table*}

\subsection{Main Results}

Table~\ref{tab:reward_model_evaluation} summarizes reward-model performance. On the 27B teacher, \teaname achieves the best results on all metrics, improving over GRPO in both score calibration (PLCC/SRCC) and pairwise preference accuracy. On the 9B model, \stuname is also consistently best and reaches 0.8864 HPA and 0.9801 margin HPA, close to the 27B \teaname teacher. This suggests that the teacher's reasoning-conditioned score distribution can be effectively internalized into a smaller direct-scoring model.

RewardDance shows a useful contrast on 9B: compared with SFT, it improves HPA from 0.7459 to 0.7817, but decreases PLCC from 0.5296 to 0.5182 and SRCC from 0.4942 to 0.4338. This suggests that post-hoc pseudo reasoning helps the small model recognize coarse pairwise preference directions, but does not guarantee calibrated rubric scores. GRPO also trails RewardDance on 9B HPA, likely because pure self-exploration is bounded by the weaker reasoning ability of the 9B model. In contrast, \teaname provides direct distribution and score-gap supervision during exploration, giving the policy clearer optimization directions, while \stuname further uses KL supervision from the 27B teacher distribution to internalize fine-grained reasoning-based scoring behavior.

\subsection{Ablation Studies}

\textbf{Effect of decoding from score distribution instead of parsing text to compute rewards.}
We compare two ways of extracting rewards from a generative reward model. The first follows common generative reward modeling practice: parse the final textual score from the model response and use it to compute the RL reward. The second uses our score decoder to obtain the full score distribution and computes the reward from its expectation. All other training settings are kept the same for GRPO and \teaname.

\begin{figure}[t]
    \centering
    \begin{subfigure}[t]{0.24\textwidth}
        \centering
        \includegraphics[width=\linewidth]{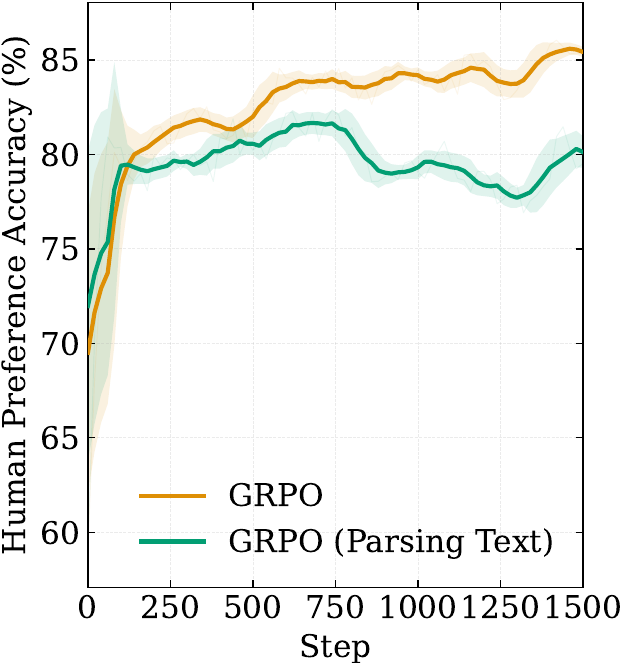}
        \caption{GRPO HPA}
        \label{fig:ab-grpo-hpa}
    \end{subfigure}\hfill
    \begin{subfigure}[t]{0.24\textwidth}
        \centering
        \includegraphics[width=\linewidth]{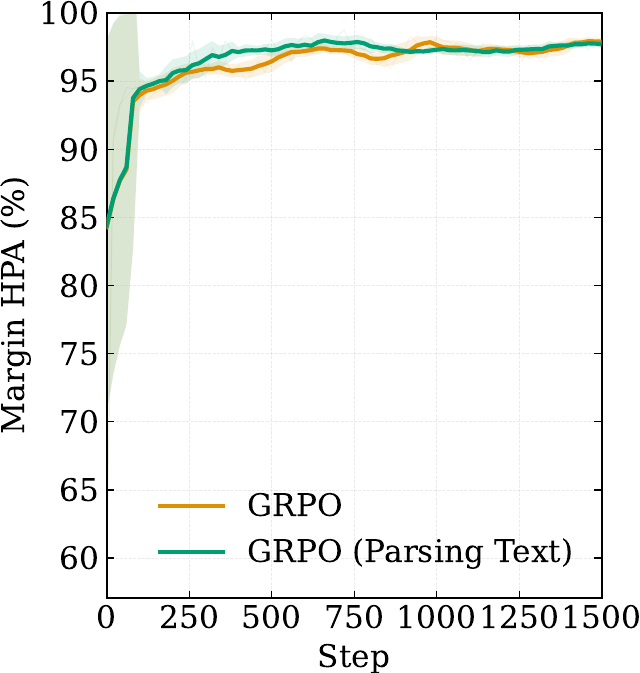}
        \caption{GRPO Margin HPA}
        \label{fig:ab-grpo-mhpa}
    \end{subfigure}\hfill
    \begin{subfigure}[t]{0.24\textwidth}
        \centering
        \includegraphics[width=\linewidth]{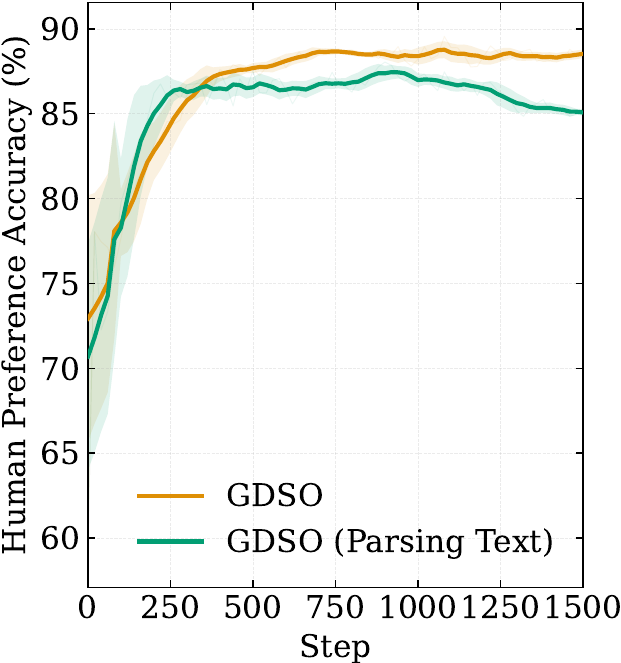}
        \caption{\teaname HPA}
        \label{fig:ab-gdso-hpa}
    \end{subfigure}\hfill
    \begin{subfigure}[t]{0.24\textwidth}
        \centering
        \includegraphics[width=\linewidth]{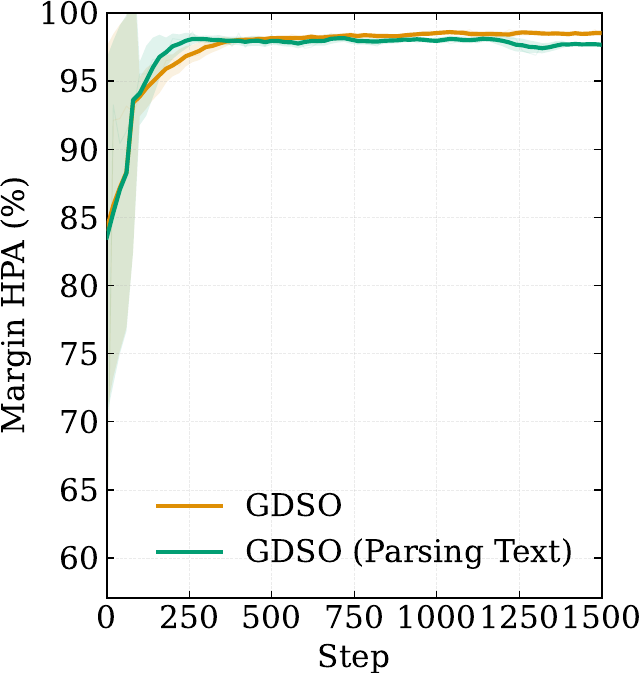}
        \caption{\teaname Margin HPA}
        \label{fig:ab-gdso-mhpa}
    \end{subfigure}
    \caption{\textbf{Effect of reward computation from score distributions.} We compare human preference accuracy and margin human preference accuracy when rewards are computed from decoded score distributions instead of parsed score text. ``Parsing Text'' denotes computing the reward from the score parsed from the generated text, rather than from the expectation of the score distribution.}
    \label{fig:ab-score-distribution-decoding}
\end{figure}

As shown in~\figref{fig:ab-score-distribution-decoding}, using the distribution expectation consistently improves both HPA and margin HPA for GRPO and \teaname. Parsing text scores effectively quantizes the reward signal: predictions such as 3.8 and 4.2 may both be emitted as the score token 4, so they receive the same reward and the same normalized advantage in GRPO. This removes fine-grained scoring signals and slows reward-model learning. In contrast, the expectation over the score distribution preserves uncertainty across neighboring bins, providing denser supervision and better teaching the model how to score.

\textbf{Distill Reasoning Chains from Teacher to Student via On-Policy Distillation.}
We compare \stuname with a direct on-policy distillation (OPD)~\cite{lu2025onpolicy} baseline, following the per-token reverse-KL formulation used by Thinking Machines Lab and related reverse-KL distillation objectives for generative LMs~\cite{gu2024minillm}. In OPD, the 9B student first samples its own reasoning-and-score trajectory $y=(y_1,\ldots,y_T)\sim\pi_{\phi_{\mathrm{old}}}(\cdot\mid x)$ for an input $x=(p,I,d)$, and the 27B \teaname teacher only provides token log-probabilities on the student's visited prefixes. The per-token reverse-KL term is used as an on-policy advantage:
\begin{equation}
A^{\mathrm{OPD}}_t
=
\log \pi_{\theta_T}(y_t\mid y_{<t},x)
-
\log \pi_{\phi_{\mathrm{old}}}(y_t\mid y_{<t},x).
\label{eq:opd-advantage}
\end{equation}
The student is then updated by an on-policy policy-gradient objective, with $A^{\mathrm{OPD}}_t$ treated as a stop-gradient advantage:
\begin{equation}
\mathcal{L}_{\mathrm{OPD}}
=
-\mathbb{E}_{x,\;y\sim\pi_{\phi_{\mathrm{old}}}(\cdot\mid x)}
\left[
\frac{1}{T}\sum_{t=1}^{T}
\operatorname{sg}\!\left(A^{\mathrm{OPD}}_t\right)
\log \pi_\phi(y_t\mid y_{<t},x)
\right].
\label{eq:opd-objective}
\end{equation}

\begin{wraptable}{r}{0.5\textwidth}
  \vspace{-0.6em}
  \centering
  \caption{\textbf{Trajectory- vs. outcome-level distillation.} HPA denotes human preference accuracy, and output tokens denote generated output length. \textbf{Best} and \underline{second-best} results are highlighted.}
  \label{tab:opd_human_preference_accuracy}
  \small
  \setlength{\tabcolsep}{0pt}
  \begin{tabularx}{\linewidth}{@{}>{\hsize=1.25\hsize\centering\arraybackslash}X|>{\hsize=0.85\hsize\centering\arraybackslash}X>{\hsize=0.95\hsize\centering\arraybackslash}X>{\hsize=0.95\hsize\centering\arraybackslash}X@{}}
  \toprule
  \multirow{2}{*}{\textbf{Method}} & \multirow{2}{*}{\textbf{HPA}} & \textbf{Margin} & \textbf{Output} \\
  & & \textbf{HPA} & \textbf{Tokens} \\
  \midrule
  9B OPD & 0.8311 & \underline{0.9643} & {\textasciitilde}750 \\
  \midrule
  9B SFT & 0.7459 & 0.8401 & \textbf{1} \\
  9B \teaname & \underline{0.8395} & 0.9599 & {\textasciitilde}750 \\
  \rowcolor[RGB]{240,230,245}[0pt][0pt]
  9B \stuname & \textbf{0.8864} & \textbf{0.9801} & \textbf{1} \\
  \bottomrule
  \end{tabularx}
  \vspace{-0.6em}
\end{wraptable}

Table~\ref{tab:opd_human_preference_accuracy} shows that OPD improves over SFT and reaches a similar level to the 9B \teaname model, but it still does not approach the 27B teacher or the 9B \stuname student.
The output-token column further reveals an efficiency gap: OPD and \teaname require long autoregressive reasoning traces, averaging about 750 output tokens, while \stuname returns the score in a single output token, matching SFT's decoding cost. Since reward inference is repeatedly called during candidate filtering or optimization, this reduction directly lowers latency and serving cost. Thus, \stuname does not merely improve HPA; it transfers the teacher's reasoning benefit into a deployment-efficient outcome-level scorer. The remaining accuracy gap suggests that teacher-derived token advantages are not sufficient when the 9B student cannot explore strong reasoning trajectories by itself: OPD can reinforce better tokens on the student's on-policy prefixes, but its learning signal is still bounded by the states the student visits. OPD and \teaname almost reach a similar 9B ceiling by giving direct score-distribution and score-gap guidance during exploration. In contrast, \stuname provides a finer-grained supervision signal by directly matching the teacher's reasoning-conditioned score distribution over the score vocabulary. This allows the student to internalize the teacher's scoring behavior without having to reproduce the full reasoning trajectory through on-policy exploration.

\section{Validating \mname as an Optimizable Reward Signal}
\label{sec:reward_feedback_learning}

To demonstrate the practical utility of \mname, we apply it to the Reinforcement Learning (RL) stage of text-to-image generation, a setting where prior work has explored policy-gradient fine-tuning, direct preference optimization, and differentiable reward backpropagation~\cite{fan2023dpok,wallace2024diffusion,prabhudesai2024alignprop,clark2024directlyfinetuningdiffusionmodels}. Unlike traditional discrete scalar rewards that provide sparse guidance, the score distributions predicted by \mname offer dense and informative gradient signals. We leverage these gradients to directly optimize the baseline SFT model~\cite{Team2025ZImageAE}, steering the generation toward human preferences.

\subsection{Multi-Dimensional Reward Gradient Backpropagation}

\begin{figure}[t]
    \centering
    \begin{subfigure}[t]{0.48\textwidth}
        \centering
        \includegraphics[width=\linewidth]{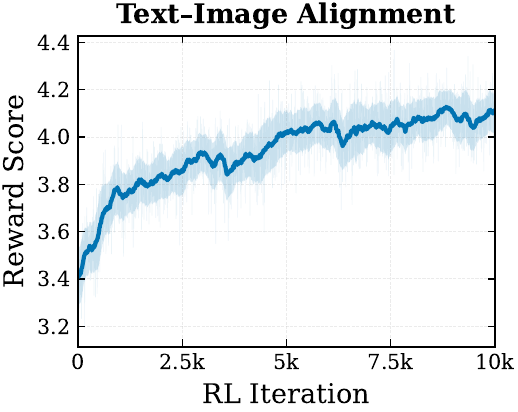}
        \caption{Text--Image Alignment}
        \label{fig:rl_alignment}
    \end{subfigure}
    \hfill
    \begin{subfigure}[t]{0.48\textwidth}
        \centering
        \includegraphics[width=\linewidth]{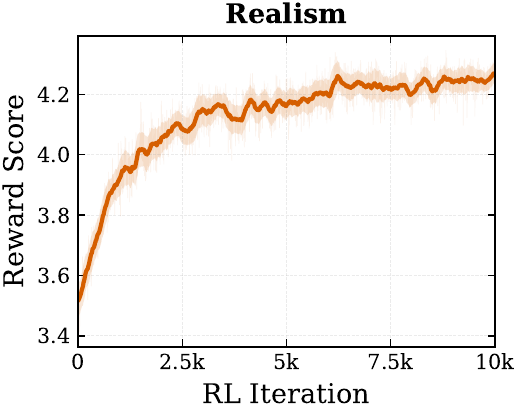}
        \caption{Realism}
        \label{fig:rl_realism}
    \end{subfigure}

    \vspace{0.8em}

    \begin{subfigure}[t]{0.48\textwidth}
        \centering
        \includegraphics[width=\linewidth]{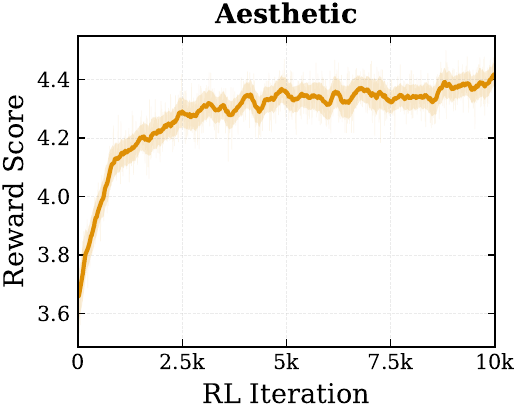}
        \caption{Aesthetics}
        \label{fig:rl_aesthetic}
    \end{subfigure}
    \hfill
    \begin{subfigure}[t]{0.48\textwidth}
        \centering
        \includegraphics[width=\linewidth]{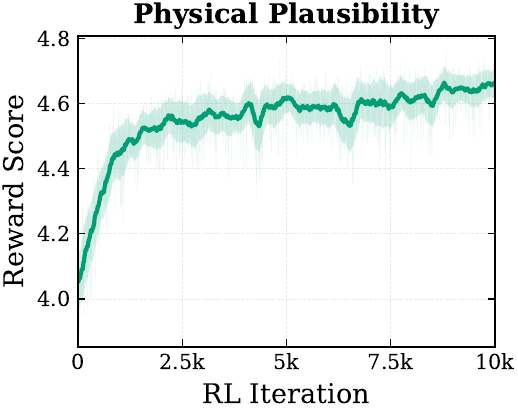}
        \caption{Physical Plausibility}
        \label{fig:rl_physics}
    \end{subfigure}
    \caption{Validation reward trajectories during RL-based text-to-image optimization using \mname. We report reward score trends over 10k RL iterations for four optimized dimensions: text--image alignment, realism, aesthetics, and physical plausibility. All scores are reported on the five-level rubric scale, where higher is better.}
    \label{fig:rl_reward_curves}
\end{figure}

We adopt a ReFL-style~\cite{xu2023imagereward} direct reward backpropagation scheme, extended to earlier denoising steps and multi-dimensional reward optimization, which is closely related to differentiable reward fine-tuning and dense reward views of the diffusion trajectory~\cite{clark2024directlyfinetuningdiffusionmodels,prabhudesai2024alignprop,yang2024adensereward}. Given a prompt $p$ and a generated image $I = G_\psi(p)$, the deployed student reward model predicts $q_\phi(s \mid p,I,d)$ for each reward dimension $d \in \mathcal{D}$, where $\mathcal{D}$ includes text--image alignment, realism, aesthetics, and physical plausibility. We use the expected score $\mu_\phi(p,I,d)$ defined in Eq.~\ref{eq:student-expected-score} as the reward for dimension $d$, and aggregate the multi-dimensional rewards as:
\begin{equation}
R(p,I) = \mathcal{A}\big(\{\mu_\phi(p,I,d)\}_{d \in \mathcal{D}}\big),
\label{eq:aggregate-reward}
\end{equation}
where $\mathcal{A}(\cdot)$ denotes a task-dependent aggregation function. We then backpropagate $\nabla_\psi R(p,G_\psi(p))$ through the denoising process to update the generator.

\begin{figure*}[t]
  \centering
  \includegraphics[width=\textwidth]{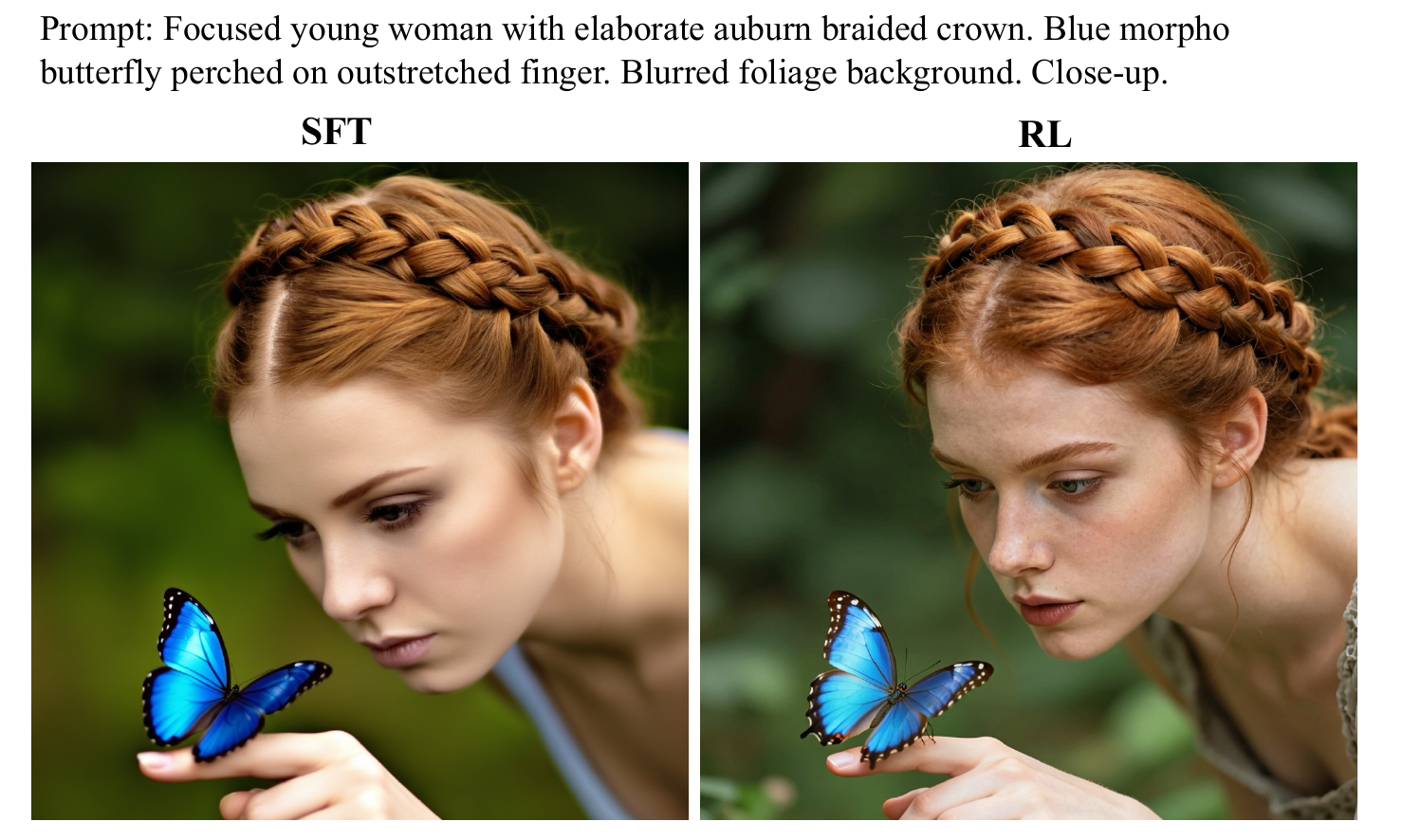}
  \vspace{0.4em}
  
  \includegraphics[width=\textwidth]{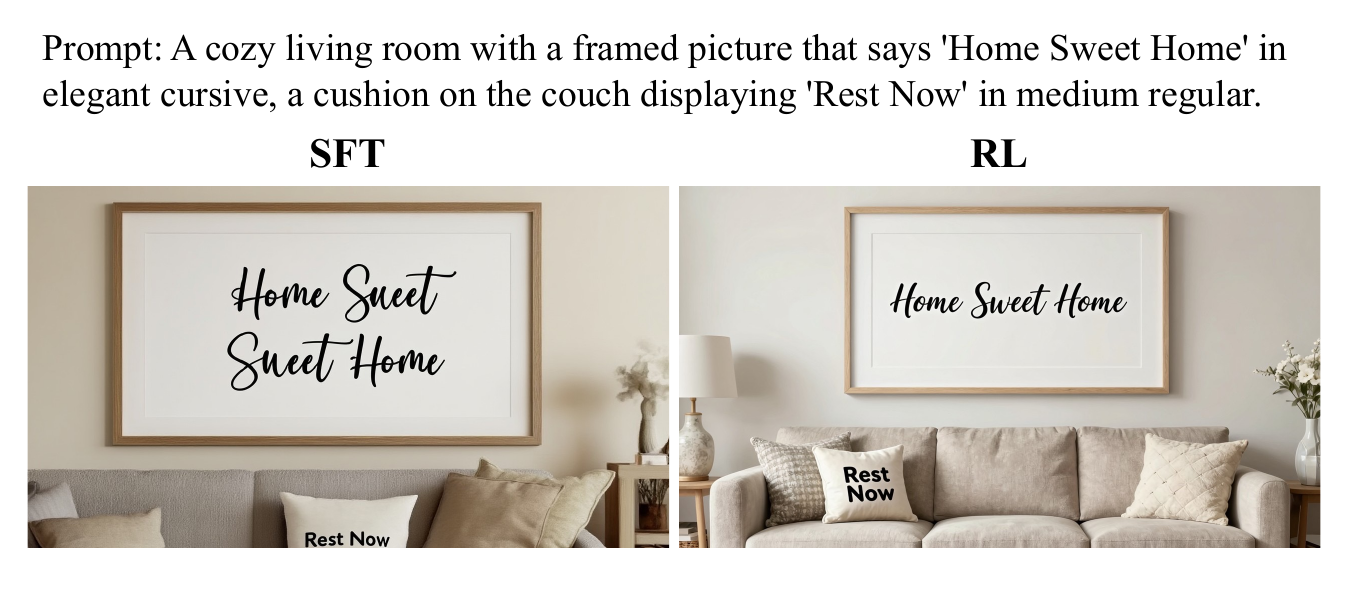}
  
  \caption{\textbf{Qualitative comparisons between the SFT baseline and \mname-guided optimization.}
  Each row shows one held-out prompt and compares the baseline generation with the optimized model.}
  \label{fig:qual_page1}
\end{figure*}

\subsection{Reward Curve Analysis}
Figure~\ref{fig:rl_reward_curves} shows that reward-guided optimization steadily improves validation rewards across text--image alignment, realism, aesthetics, and physical plausibility. Realism and aesthetics improve faster in the early stage, while text--image alignment and physical plausibility increase more gradually due to their stronger dependence on semantic and structural correctness. These trends suggest that \mname provides stable and fine-grained optimization signals across multiple aspects of visual generation.

\subsection{Human Evaluation}
To examine whether the reward improvements translate into human-perceived quality gains, we conduct blind human evaluation on the same held-out prompt set used in the validation reward analysis, following the broader emphasis on reproducible human evaluation for text-to-image generation~\cite{otani2023toward}. The set contains 400 prompts covering compositional descriptions, attribute binding, spatial relations, and physically challenging scenes, matching the compositional coverage studied in recent T2I and text-to-visual evaluation benchmarks~\cite{hu2023tifa,ghosh2023geneval,huang2023t2icompbench,lin2024evaluating,li2024evaluating}. Professional annotators perform pairwise comparisons between images generated by the SFT baseline and those generated by the model optimized with \mname.


We report results using the human-preference-based \textbf{Good-Same-Bad (GSB)} metric. For each prompt, annotators judge whether the optimized image is better than, comparable to, or worse than the baseline image, corresponding to \textit{Good}, \textit{Same}, and \textit{Bad}, respectively. Let $G$, $S$, and $B$ denote the counts of these three outcomes. The final GSB score is defined as
\begin{equation}
    GSB = \frac{G - B}{G + S + B}.
    \label{eq:gsb}
\end{equation}
where a higher value indicates stronger net human preference for the optimized model.

Compared with the strong SFT baseline, the model optimized using \mname achieves a net GSB improvement of \textbf{41.3\%}. This result confirms that the improvements measured by our reward model are reflected in human judgments, which is important because reward-guided optimization can otherwise overfit proxy rewards~\cite{gao2023scaling,rafailov2024scaling}. Qualitative comparisons in Figure~\ref{fig:qual_page1} further illustrate that \mname-guided optimization improves text-image alignment, visual realism, aesthetics, and physical plausibility across diverse prompts.

\section{Discussions and Future Works}

\textbf{Reasoning-score coupling.}
One limitation of the current teacher training objective is that it combines policy-gradient rewards with direct SFT-style losses, $\alpha_{\mathrm{pt}}\mathcal{L}^{\mathrm{pt}}_{\mathrm{CE}}+\alpha_{\mathrm{pw}}\mathcal{L}^{\mathrm{pw}}$ in Eq.~(\ref{eq:gdso-objective}). These supervised terms substantially improve score calibration and preference metrics, but they may occasionally make the final score depend more on direct score supervision than on the generated reasoning trace itself. In our experiments, such weak coupling between reasoning and the final score appears to be a minority case, while the metric gains from these losses are clear. Future work could add explicit reasoning-score consistency checks or contrastive supervision so that the teacher preserves the calibration benefits of direct losses while keeping the score more tightly grounded in its rationale.

\textbf{Potential Generalization to All Sequence-to-Score Tasks.}
Although this paper focuses on reward modeling for image generation, the proposed formulation is not tied to a specific visual domain. As VLMs continue to improve, a VLM-to-score model can naturally take image, video, or text-centered inputs and convert arbitrary model outputs into rubric-aligned score distributions. In our framework, the score can be decoded during the reasoning process or from any scoring-oriented output of the VLM, and the distribution expectation remains directly supervised and differentiable. This makes the reward useful not only as an evaluator, but also as a continuous optimization signal that can be backpropagated to generated images or videos through a differentiable generator. Beyond visual generation, the same decoupled teacher-student design can serve as a reward model for LLMs and VLMs, or as a general sequence-to-score evaluator for tasks such as image/video quality assessment and caption evaluation, connecting to broader evaluation lines such as reference-free caption metrics and fine-grained video-generation benchmarks~\cite{hessel2021clipscore,huang2024vbench}. Our experiments instantiate this idea on text-to-image generation, while broader modalities and downstream reward-modeling settings remain promising future directions.

\textbf{Possible Extension to Unified Reward Modeling.}
The same formulation also points toward unified reward modeling. Our annotation pipeline already provides more than isolated pointwise labels: when annotators score multiple candidates under the same prompt, their adjustments implicitly encode comparison signals among candidates. This enables comparison training, but with richer supervision than binary preferences. Because each candidate has a calibrated rubric score, the model can learn not only whether one sample is better than another, but also by how many score levels they differ. Such score-gap supervision is naturally compatible with our pointwise distribution objective and can be extended across dimensions, modalities, and task types. A future unified reward model could therefore combine pointwise score distributions, pairwise preferences, and calibrated score gaps within one teacher-student system, using the teacher for reasoning-heavy judgment and the student for efficient direct scoring.

\section{Related Work}
\subsection{Reward Models}

Reward models are widely used to align generative models with human preferences. Early visual reward models, such as ImageReward~\cite{xu2023imagereward}, PickScore~\cite{kirstain2023pickapicopendatasetuser}, and HPSv2~\cite{wu2023humanpreferencescorev2}, are mostly built on CLIP-style encoders and trained to output scalar preference scores. Recent VLM-based reward models further improve visual understanding and task coverage by replacing CLIP encoders with stronger multimodal backbones and attaching regressive reward heads, including VisionReward~\cite{xu2026visionrewardfinegrainedmultidimensionalhuman}, VideoAlign~\cite{liu2025improvingvideogenerationhuman}, HPSv3~\cite{ma2025hpsv3widespectrumhumanpreference}, and WorldPM~\cite{wang2025worldpmscalinghumanpreference}. Scalar or regressive reward models are efficient and convenient for deployment, but they can over-compress subjective preferences and may be vulnerable to reward hacking or reward overoptimization~\cite{you2025teachinglargelanguagemodels,wu2025rewarddancerewardscalingvisual,gao2023scaling,rafailov2024scaling}. This motivates reward modeling paradigms that preserve richer judgment information while remaining usable for optimization. 

Generative reward models aim to better exploit the native next-token prediction and reasoning capabilities of VLMs, following the broader trend of LLM/VLM-as-a-judge systems~\cite{zheng2023judging,gu2024surveyllmasajudge,chen2024mllmjudge,chen2024mjbench}. Representative works include DeepSeek-GRM~\cite{liu2025inferencetimescalinggeneralistreward}, GenRM-CoT~\cite{zhang2025generativeverifiersrewardmodeling}, UnifiedReward~\cite{wang2026unifiedrewardmodelmultimodal}, RewardDance~\cite{wu2025rewarddancerewardscalingvisual}, and Edit-R1~\cite{guo2026leveraging}. RewardDance formulates reward prediction as a generative comparison task and studies scaling along model and context dimensions. Edit-R1 further shows that verifier-style reasoning, which decomposes editing instructions into explicit principles and verifies outputs with CoT, can provide stronger feedback for image editing. Orthogonally, score-distribution modeling has been explored in quality assessment and ordinal label learning~\cite{murray2012ava,talebi2018nima,diaz2019soft,wen2023ordinal}. Q-Align~\cite{wu2023qalign} discretizes continuous scores into level tokens, while DeQA~\cite{you2025teachinglargelanguagemodels} shows that distribution-based soft labels better preserve uncertainty and inter-image relationships than one-hot labels. Different from these works, \mname uses a reasoning VLM to infer rubric-aligned score distributions and further distills them into an efficient direct-scoring student, thereby combining reasoning-aware judgment with deployable distributional rewards.

\subsection{Reinforcement Learning for Visual Generation}
Reinforcement learning from human feedback has been increasingly used to align visual generators with human preferences. Existing methods either adapt policy-gradient algorithms to diffusion or flow-based generators, such as DDPO~\cite{black2024trainingdiffusionmodelsreinforcement}, DPOK~\cite{fan2023dpok}, and recent GRPO-style visual RL methods~\cite{deepseek-math,liu2025improvingvideogenerationhuman,wu2025rewarddancerewardscalingvisual,guo2026leveraging}, optimize diffusion models from pairwise preferences~\cite{rafailov2023direct,wallace2024diffusion}, or directly backpropagate reward gradients through the sampling process, like ReFL~\cite{xu2023imagereward}, DRaFT~\cite{clark2024directlyfinetuningdiffusionmodels}, AlignProp~\cite{prabhudesai2024alignprop}, and dense reward formulations~\cite{yang2024adensereward}. Related preference-optimization and online RL methods have further been explored for text-to-image generation, video generation, and image editing~\cite{liu2025improvingvideogenerationhuman,wu2025rewarddancerewardscalingvisual,guo2026leveraging}. These works show that reward-guided optimization can substantially improve visual generation quality, but its effectiveness depends heavily on the reward signal. Scalar or regressive rewards are efficient and convenient for optimization, but they can over-compress subjective preferences and may be vulnerable to reward hacking~\cite{you2025teachinglargelanguagemodels,guo2026leveraging,gao2023scaling,rafailov2024scaling}. Reasoning-based rewards provide richer semantic verification, but explicit reasoning traces can introduce additional inference overhead or become incompatible with direct reward backpropagation~\cite{yang2026joint,guo2026leveraging}. In contrast, \mname distills reasoning-enhanced judgments into score distributions whose expectations provide dense and differentiable rewards, enabling efficient reward-guided optimization of text-to-image generators.

\subsection{On-policy Distillation}
A practical reward model should be efficient enough for large-scale scoring and sufficiently stable for reward-guided optimization, since visual reward models are commonly used as automatic evaluators or optimization signals for improving generated samples~\cite{xu2023imagereward,wu2025rewarddancerewardscalingvisual,yang2026joint}.
On-policy distillation (OPD) has recently emerged as a relevant paradigm for transferring reasoning behaviors from stronger models to weaker ones. Instead of relying only on fixed offline trajectories, OPD trains the student on its own sampled trajectories and uses a teacher to provide dense supervision on the states visited by the student~\cite{Agarwal2023OnPolicyDO,lu2025onpolicy}. This makes the learning signal better matched to the student's inference-time distribution and is particularly relevant for long-horizon reasoning, where deviations from offline traces can accumulate over multiple steps. Recent studies have extended this idea to self-distillation, privileged-information distillation, reasoning compression, continual learning, and reward-to-supervision conversion~\cite{zhao2026selfdistilledreasoneronpolicyselfdistillation,penaloza2026privileged,shenfeld2026selfdistillationenablescontinuallearning,Sang2026CRISPCR,he2026selfdistillationzeroselfrevisionturns}. Follow-up work further analyzes the failure modes, stability issues, and practical design choices of OPD in reasoning distillation~\cite{song2026surveyonpolicydistillationlarge,fu2026revisitingonpolicydistillationempirical,li2026rethinkingonpolicydistillationlarge,jang2026stableonpolicydistillationadaptive,zhu2026facesonpolicydistillationpitfalls}.


While OPD provides a natural reference for transferring a large reasoning teacher into a compact student, its trajectory-centric objective differs from the deployment goal of visual reward modeling: in reward-guided generation, the reward model is commonly used as an efficient scorer or a differentiable optimization signal for selecting or directly improving generated samples~\cite{xu2023imagereward,clark2024directlyfinetuningdiffusionmodels,yang2026joint}, whereas reasoning-based reward models that generate discrete multi-step traces can be costly or incompatible with direct reward backpropagation~\cite{guo2026leveraging}. These observations suggest that directly applying OPD to distill reasoning trajectories is not the most natural objective for visual reward modeling: it teaches the student how the teacher reasons, while deployment only requires how the teacher judges. In contrast, \mname distills the outcome of reasoning. The teacher first uses reasoning to produce a calibrated score distribution, and the student learns to predict this distribution directly. This reasoning-internalized distillation transfers the teacher's judgment behavior while avoiding explicit reasoning at inference time.

\section{Conclusion}

We presented \mname, a teacher-student framework for visual reward modeling that represents human preference as a reasoning-conditioned score distribution rather than a single deterministic scalar. By training a large VLM teacher with Group-wise Direct Score Optimization, \mname combines policy-gradient learning with direct supervision on score distributions and score gaps, improving both calibrated scoring and pairwise preference ranking. Through Reasoning-Internalized Score Distillation, a compact student internalizes the teacher's reasoning-based distribution and provides efficient, direct, and differentiable scoring without generating explicit reasoning chains.

Experiments on our internally annotated benchmark show that the 27B \teaname teacher outperforms SFT, RewardDance, and GRPO, while the 9B \stuname student closely matches the larger teacher and serves as an effective reward signal for text-to-image optimization. Beyond the current image-generation setting, we view \mname as a general sequence-to-score modeling paradigm: future work can extend the same decoupled teacher-student design to unified reward modeling across image, video, text, and multimodal generation tasks, combining pointwise score distributions, pairwise comparisons, and calibrated score-gap supervision in a single reward model.

\clearpage
\bibliographystyle{splncs04}
\bibliography{main}

\end{document}